\pdfoutput=1
\documentclass[sigconf]{acmart}
\usepackage{algorithm}
\usepackage{algorithmic}
\usepackage{amsmath}
\usepackage{amssymb}
\usepackage{multirow} 
\usepackage{amsthm}
\usepackage{graphicx}

\usepackage{url}

\usepackage[utf8x]{inputenc} 
\usepackage{ragged2e}
\AtBeginDocument{%
  \providecommand\BibTeX{{%
    \normalfont B\kern-0.5em{\scshape i\kern-0.25em b}\kern-0.8em\TeX}}}
 
\usepackage{subfigure}
\AtBeginDocument{%
  \providecommand\BibTeX{{%
    \normalfont B\kern-0.5em{\scshape i\kern-0.25em b}\kern-0.8em\TeX}}}

\copyrightyear{2020} 
\acmYear{2020} 
\setcopyright{acmlicensed}\acmConference[MM '20]{Proceedings of the 28th ACM International Conference on Multimedia}{October 12--16, 2020}{Seattle, WA, USA}
\acmBooktitle{Proceedings of the 28th ACM International Conference on Multimedia (MM '20), October 12--16, 2020, Seattle, WA, USA}
\acmPrice{15.00}
\acmDOI{10.1145/3394171.3413597}
\acmISBN{978-1-4503-7988-5/20/10}
\begin{document}
\fancyhead{}
\title[Subspace Clustering via Graph Filtering]{Towards Clustering-friendly Representations: Subspace Clustering via Graph Filtering}

\author{Zhengrui Ma}
\author{Zhao Kang}\authornote{Corresponding author}
\affiliation{
  \institution{School of Computer Science and Engineering, University of Electronic Science and Technology of China}
  }
  \email{zkang@uestc.edu.cn}
\author{Guangchun Luo}
  \affiliation{
  \institution{School of Information and Software Engineering, University of Electronic Science and Technology of China}
  }
 \email{gcluo@uestc.edu.cn}
  \author{Ling Tian, Wenyu Chen}
  \affiliation{
  \institution{School of Computer Science and Engineering, University of Electronic Science and Technology of China}
  }
 \email{{lingtian,cwy}@uestc.edu.cn}
\renewcommand{\shortauthors}{}


\begin{abstract}
Finding a suitable data representation for a specific task has been shown to be crucial in many applications. The success of subspace clustering depends on the assumption that the data can be separated into different subspaces. However, this simple assumption does not always hold since the raw data might not be separable into subspaces. To recover the ``clustering-friendly'' representation and facilitate the subsequent clustering, we propose a graph filtering approach by which a smooth representation is achieved. Specifically, it injects graph similarity into data features by applying a low-pass filter to extract useful data representations for clustering. Extensive experiments on image and document clustering datasets demonstrate that our method improves upon state-of-the-art subspace clustering techniques. Especially, its comparable performance with deep learning methods emphasizes the effectiveness of the simple graph filtering scheme for many real-world applications. An ablation study shows that graph filtering can remove noise, preserve structure in the image, and increase the separability of classes.
\end{abstract}




\maketitle

\section{Introduction}
Clustering is a long-standing problem in machine learning, data mining, and pattern recognition, with an endless of applications. It is also a notoriously hard task due to its unsupervised learning nature \cite{zhang2019flexible,kang2020structure,peng2017nonnegative}. Its performance can be easily affected by many factors, such as data representation, feature dimension, and noise \cite{hu2017robust,kang2020relation,ertoz2003finding,kang2020robust}. Among various clustering techniques, K-means and spectral clustering are especially popular.

K-means is suitable for data that are evenly spread around some centroids \cite{lloyd1982least,peng2018integrate,liu2020efficient}. In many real-life applications, the data might not be separable. A number of techniques, including kernel trick, principal component analysis, canonical correlation analysis, have been developed to map high-dimensional data to a certain representation that is suitable for performing K-means. Spectral clustering is basically a generalization of kernel K-means \cite{ng2002spectral,wen2020generalized}. They provide meaningful outputs only when the data are mapped to a ``clustering-friendly'' representation, in which the data samples nicely fall into clusters. 

To tackle the curse of dimensionality, subspace clustering (SC) assumes that data lie in a union of subspaces \cite{vidal2011subspace,liu2019robust}. SC has well-documented impact in a wide range of applications \cite{zhang2020generalized}. It has been pointed out that applying subspace clustering on the projected data is beneficial since the original data might not fall on separate subspaces \cite{liu2011latent,patel2015latent}.

In this paper, instead of applying subspace clustering on the original space, we learn the subspace clustering in a ``clustering-friendly'' representation, which is easy to cluster. Even if the data cannot be separated in the original domain, its smooth representation can be grouped into disjoint subspaces. In particular, we inject graph similarity into data features by applying a low-pass filter to extract meaningful data representations for clustering. Since the graph is unavailable beforehand, an iterative approach is used. The proposed framework can incorporate various subspace clustering models. 

Our contributions are summarized as follows.
\begin{itemize}
    \item We propose a graph filtering framework for subspace clustering, which generates a ``clustering-friendly'' representation. This provides a new representation learning strategy.
    \item Taking two representative subspace clustering techniques as examples, we demonstrate the considerable enhancement brought by graph filtering on a number of datasets. 
    \item Graph filtering approach produces comparable results with respect to state-of-the-art deep neural networks based clustering techniques.
    \item An ablation study shows that graph filtering can remove noise, preserve structure in the image, and increase the separability of classes.
\end{itemize}
\section{Graph Filtering}

Suppose an undirected graph $G=(\mathcal{V},W,X)$ with $n=|\mathcal{V}|$ vertices is given, with an edge weights matrix $W\in\mathcal{R}^{n\times n}$, where $w_{ij} = w_{ji} \geq 0$, and a feature matrix $X=[\emph{\textbf{x}}_1,\cdots,\emph{\textbf{x}}_n]^\top\in\mathcal{R}^{n\times m}$ corresponding to $n$ vertices. The degree of vertex $v_{i}$ is defined as $D_{ii} = \sum_{j=1}^{n}w_{ij}$ and $D=diag(d_1,\cdots,d_n)$. The symmetrically normalized graph Laplacian $L_{s} = I - D^{\frac{1}{2}}WD^{\frac{1}{2}}$ can be eigen-decomposed as $L_{s} = U\Lambda U^{-1}$, where the associated eigenvalues $\Lambda=diag(\lambda_1,\cdots, \lambda_n)$ are sorted in increasing order and $U=[\emph{\textbf{u}}_1,\cdots,\emph{\textbf{u}}_n]$ are the corresponding orthogonal eigenvectors. The set of eigenvectors of $L_{s}$ can be considered as Fourier basis of the graph and the eigenvalues $\lambda_{i}$ can be considered as the associated frequencies \cite{shuman2013emerging}.\par
Let $ f : \mathcal{V} \rightarrow \mathbb{R}$ be a real-valued function on the nodes of a graph, a \emph{graph signal} $ \emph{\textbf{f}} = [f(v_{1}),f(v_{2}),...,f(v_{n})]^{\top}$ can be represented as a linear combination of the eigenvectors, i.e.,
\begin{equation}
\emph{\textbf{f}} = \sum_{i=1}^{n}c_{i}\emph{\textbf{u}}_{i} = U\emph{\textbf{c}},
\end{equation}
where $\emph{\textbf{c}}= [c_{1},c_2,...,c_n]^\top$ is the coefficient vector. The absolute value of $c_i$ shows the strength of the basis signal $\emph{\textbf{u}}_{i}$ presented in graph signal \emph{\textbf{f}}.
The smoothness of \emph{\textbf{f}} can be measured by
\begin{equation}
\begin{aligned}
E_{f} &= \frac{1}{2}\sum_{i,j=1}^{n}w_{ij}\|\frac{{f}_{i}}{\sqrt{d_{i}}}-\frac{{f}_{j}}{\sqrt {d_{j}}}\|_2^{2} =  {\emph{\textbf{f}}}^{\top} L_{s} \emph{\textbf{f}}\\
      &= {(U\emph{\textbf{c}})}^{\top}L_{s}U\emph{\textbf{c}} = \sum_{i=1}^{n} {c_i}^{2}\lambda_{i}.
\end{aligned}
\end{equation}
This indicates that the basis signals corresponding to smaller $\lambda_{i}$ are smoother. Hence, a smooth signal $\emph{\textbf{f}}$ should consist of more low-frequency basis signals than high-frequency ones \cite{chung1997spectral}.\\

The graph signals associated with the real-world data should be sufficiently smooth, i.e., the signal values should change gradually across connected neighbor nodes. This can be achieved through a low-pass graph filter $G$. 
%We can learn the smooth feature matrix $\bar{X}$ via graph convolution with the weighted matrix $W$  and its Laplacian $L_{s}$ obtained from subspace segmentation. 
Assume $h(\lambda_{i})$ is a low-pass frequency response function, the filtered signal $\bar{\emph{\textbf{f}}}$ can be written as
\begin{equation}
\bar{\emph{\textbf{f}}} =G\emph{\textbf{f}}= \sum_{i=1}^{n}h(\lambda_{i})c_{i}\emph{\textbf{u}}_{i} = UH(\Lambda)\emph{\textbf{c}} = UH(\Lambda)U^{-1}\emph{\textbf{f}},
\end{equation}
where $H(\Lambda)=diag(h(\lambda_{1}),h(\lambda_{2}),...,h(\lambda_{n}))$. To preserve the low-frequency signals and remove the high-frequency ones, $h(\lambda_{i})$ should be large for small $\lambda_{i}$ and vice versa. Since the eigenvalues of symmetrically normalized graph Laplacian $L_s$ fall to range $[0,2]$, one choice of the low-pass response function is $h(\lambda_{i}) = (1 - \frac{\lambda_{i}}{2})^{k}$, where positive integer $k$ is applied to capture the $k$-hop neighborhood relations \cite{chung1997spectral,zhang2019attributed}. Then the filtered signal can be formulated as
\begin{equation}
\bar{\emph{\textbf{f}}} =  U(I-\frac{\Lambda}{2})^{k}U^{-1}\emph{\textbf{f}} = (I-\frac{L_s}{2})^{k}\emph{\textbf{f}}.
\label{filterequation}
\end{equation}
Each column of $X$ can be taken as a graph signal. Then, a smoothed representation $\bar{X}$ is achieved by
\begin{equation}
    \bar{X}=(I-\frac{L_s}{2})^{k}X.
    \label{xbar}
\end{equation}
In essence, $\bar{x}_i$ is obtained by aggregating the features of its $k$-hop neighbors iteratively. Thus, a $k$-order graph filtering takes into account long-distance data relations, which would be useful for capturing global structure to improve downstream task performance.

\section{The Proposed Methodology}
Samples drawn from the same cluster tend to be densely connected, thus it is natural to assume that they are likely to have similar feature representations \cite{wang2020smooth}. To this end, we can obtain ``clustering-friendly'' representations by using graph filtering. %smoothness of feature matrix is a desired property.

Subspace clustering constructs an affinity graph matrix $W$ from feature matrix $X$ for the subsequent spectral clustering task. In this work, we aim to learn $W$ in a smooth representation $\bar{X}$. However, to compute $\bar{X}$, we need to know the affinity graph $W$ in advance. To address this dilemma, we propose an iterative approach.% unified framework to learn both affinity matrix $W$ and filtered feature $\bar{X}$ iteratively.

Our proposed graph filtering perspective can be generally integrated with various kinds of subspace clustering models. Due to its simplicity and effectiveness, Least Square Regression (LSR) is a very popular subspace clustering model as shown in Eq.(17) in \cite{lu2012robust}. Hence, we choose it to demonstrate our proposed method. Suppose we have smooth representation $\bar{X}$, LSR learns a coefficient matrix $Z$ by
\begin{equation}
\min_{Z}  \|\bar{X}^{\top}-\bar{X}^{\top}Z\|_F^2+\alpha \|Z\|_F^2.
\label{FLSR}
\end{equation}

Its closed-form solution can be achieved by setting its first-order derivative to zero, which yields
\begin{equation}
Z = (\bar{X}\bar{X}^{\top}+\alpha I)^{-1}\bar{X}\bar{X}^{\top}.
\label{csol}
\end{equation}
Since
\begin{equation}
\begin{aligned}
{Z}^{\top} &= \bar{X}\bar{X}^{\top}(\bar{X}\bar{X}^\top+\alpha I)^{-1}\\
&= (\bar{X}\bar{X}^\top+\alpha I-\alpha I)(\bar{X}\bar{X}^\top+\alpha I)^{-1}\\
&= I - \alpha(\bar{X}\bar{X}^\top+\alpha I)^{-1}\\
&= (\bar{X}{\bar{X}^\top+\alpha I)^{-1}(\bar{X}\bar{X}^\top+\alpha I-\alpha I) = Z},
\end{aligned}
\end{equation}
we can directly set $W=|Z|$, which in turn can be used to update $\bar{X}$. To start the iterative algorithm, smooth representation $\bar{X}$ can be initialized to raw feature matrix $X$, so that an initial graph $W$ can be achieved by Eq. (\ref{csol}). We can stop the iterations when the difference between affinity matrices obtained in the $t$-th and $(t+1)$-th iteration is smaller than a threshold $\epsilon$, i.e., ${\lVert W_t-W_{t-1}\rVert}_{F}^{2}<\epsilon$. Afterwards, the spectral clustering is utilized upon $W$. The complete procedure for our algorithm is outlined in Algorithm 1. 

In Algorithm \ref{alg:algorithm}, the cost for updating $Z$ is $\mathcal{O}(\max(m,n)n^2)$. Updating $W$ takes  $\mathcal{O}(n^2)$. To compute $L$, we need $\mathcal{O}(n^2)$ time. The cost for updating $\bar{X}$ is $\mathcal{O}(n^3)$. Hence, the overall complexity is $\mathcal{O}(t(\max(m,n)n^2))$, where $t$ is the number of iterations. In fact, our method can be easily modified to deal with large-scale data. For the graph filtering part, since the graph is often sparse (Let $N,d$ denote the number of nonzero entries in graph Laplacian and feature dimensions respectively), we can left multiply $X$ by $(I - \frac{L_{s}}{2})$ for $k$ times, resulting in $\mathcal{O}(Ndk)$. For subspace clustering part, we can perform it in $\mathcal{O}(n)$ time by the idea of anchor point \cite{kang2019large}. Scalability is left for future work.
%Therefore, we purpose an iterative strategy. 
%Suppose the feature matrix in $t$th iteration is known. The coefficient matrix in $t$th iteration can be learned by solving the least square regression problem\\
%\begin{equation}
%C_t = (X_{t}{X_{t}}^T+\alpha I)^{-1}X_t{X_t}^{T}
%\end{equation}

%The learned $t$th coefficient matrix is symmetric. Then the $t$th affinity matrix $W_t$ can be constructed simply by taking its absolute value and the diagnoal $D_t$ can be derived by adding each row of $W_t$. Thus, the graph Laplacian in the $t$th iteration can be written as
%\begin{equation}
%L_{t} = I - {D_{t}}^{-\frac{1}{2}}W_{t}{D_{t}}^{-\frac{1}{2}}
%\end{equation}
%With the $t$th graph Laplacian, we can achieve the $(t+1)$th filtered feature $\bar{X}_{t+1}$ by performing graph convolution on the raw feature $X$.
%\begin{equation}
%    \bar{X}_{t+1}= (I-\frac{1}{2}L_{t})^{k}X
%\end{equation}

%Approach 1:
%Use the kernel matrix $XX^T$ to compute $L$.\\
%Approach 2:
%Iteratively update $C$ and $\bar{X}$.

\begin{algorithm}[tb]
\caption{FLSR}
\label{alg:algorithm}
\begin{flushleft}
\textbf{Input}: raw feature matrix $X$\\
\textbf{Parameter}: filter order $k$, trade-off parameter $\alpha$, cluster number $g$\\
\textbf{Output}: $g$ partitions\\
\end{flushleft}
\begin{algorithmic}[1] %[1] enables line numbers

\STATE Initialize $t=0$ and $\bar{X}_1 = X$
\REPEAT
\STATE Set $t=t+1$.
    \STATE $Z_t = (\bar{X}_t {\bar{X}_t}^{\top}+\alpha I)^{-1}\bar{X}_{t}\ {\bar{X}_t}^{\top}$
    \STATE $W_t = abs(Z_t)$
    %\STATE $D_t = diag(W_t)$
    \STATE $L_{t} = I - {D_t}^{-\frac{1}{2}}W_t{D_t}^{-\frac{1}{2}}$
    \STATE $\bar{X}_{t+1}= (I-\frac{1}{2}L_{t})^{k}X_t$
    \UNTIL ${\lVert W_t-W_{t-1}\rVert}_{F}^{2} < \epsilon$
\STATE Obtain the cluster partitions by performing spectral clustering on $W$
\end{algorithmic}
\end{algorithm}

Based on LSR, Thresholding Ridge Regression (TRR) \cite{peng2015robust} was proposed later. According to the property of intra-subspace projection dominance, coefficients of two samples from one cluster in learned affinity matrix by LSR are always larger than coefficients of samples from different clusters. Therefore, an extra step can be added before spectral clustering. We can only preserve first $p$ largest values in each row of affinity matrix $W$, where the value of $p$ can be the dimensionality of subspace. For convenience, we name graph filtering based LSR and TRR as FLSR and FTRR, respectively.%Of course, this additional step is optional. Two versions of our model are tested in the next part: without thresholding step(FSC1) and with thresholding step(FSC2).

%\emph{Complexity Analysis}

\section{Experiment}
In this section, we conduct experiments to demonstrate the effectiveness of graph filtering in subspace clustering \footnote{The source code is available at https://github.com/sckangz/STRR}. 
\subsection{Dataset}
We perform clustering experiments on three face datasets (ORL, AR and Umist), two object datasets (COIL20 and COIL40), one handwritten digit dataset (MNIST) and one large scale news dataset (RCV1). Specifically, ORL is composed of 400 images with different poses and expressions from 40 individuals. AR has 840 samples from 120 subjects. Umist contains 480 images with varied poses from 20 individuals. COIL20 and COIL40 have 20 and 40 classes respectively, with each class having 72 toy images. MNIST comprises handwritten digit images of 0 to 9. We use first 100 images of each digit. There are 9625 news texts in RCV1. The statistics information is summarized in Table \ref{tab:datasets}.

\begin{table}
    \caption{Statistics of datasets.}
    \label{tab:datasets}
    \begin{tabular}{cccc}
        \toprule
        Dataset  & Samples & Classes  & Dimensions\\
        \midrule
        ORL & 400 & 40 & 1024\\
        \midrule
        AR  & 840 & 120 & 768\\
        \midrule
        Umist & 480 & 20 & 1024\\
        \midrule
        COIL20 & 1440 & 20 & 1024\\
        \midrule
        COIL40 & 2880 & 40 & 1024\\
        \midrule
        MNIST & 1000 & 10 & 784\\
        \midrule
        RCV1 & 9625 & 4 & 29992\\
        \bottomrule
    \end{tabular}
\end{table}

%To examine the performance of each model, we use the learned coefficient matrix to build the affinity matrix, as input to the subsequent spectral clustering \cite{ng2002spectral}.
\subsection{Experimental Setup}
Several representative models in subspace clustering are compared in our experiment, including Sparse Subspace Clustering (SSC) \cite{elhamifar2009sparse}, Low Rank Representation (LRR) \cite{liu2010robust}, Latent Low Rank Representation (LatLRR) \cite{liu2011latent}, Low Rank Subspace Clustering (LRSC) \cite{vidal2014low}, Least Square Regression \cite{lu2012robust}, Thresholding Ridge Regression (TRR) \cite{peng2015robust}, Sparse Subspace Clustering by Orthogonal Matching Pursuit (SSCOMP) \cite{you2016scalable}, Discriminative Unsupervised Dimensionality Reduction (DUDR) \cite{wang2015discriminative}, Scaled Simplex Representation based Subspace Clustering (SSRSC) \cite{xu2019scaled}. In particular, SSRSC was published in 2019 and has shown better performance than RSIM \cite{ji2015shape}, SMR \cite{hu2014smooth}, S3C \cite{li2017structured}, EnSC \cite{you2016oracle}, ESC \cite{you2018scalable}, etc. Hence, we do not compare with those methods. Additionally, similar to our approach, both LatLRR and DUDR perform subspace clustering in a new representation.

For fair comparison, we apply the same postprocessing step in all models. Specifically, affinity graph is constructed as the following
\begin{equation}
    W = \frac{|Z^\top|+|Z|}{2}.
\end{equation}
Then spectral clustering algorithm is implemented. Parameters in each model are well-tuned to achieve its best clustering results. For our FLSR and FTRR, we fix the threshold $\epsilon$ to $10^{-5}$ and search for proper trade-off parameter $\alpha$ and filter order $k$. For FTRR, we additionally search for a proper threshold parameter $p$, the value of which is recommended to be the dimensionality of subspace \cite{peng2015robust}.

Three popular metrics are applied to quantitatively evaluate the clustering performance. They are accuracy (ACC), normalized mutual information (NMI), and purity (PUR).\\
Accuracy is defined as\\
\begin{equation}
ACC = \frac{\sum_{i}\delta (map(l_{i})=y_{i})}{n}
\end{equation}
where $y_{i}$ and $l_{i}$ denote the ground truth label and algorithm's output of sample $i$. $\delta(\cdot)$ is the indicator function. $l_{i}$ is mapped to its best group label with Kuhn-Munkres algorithm.\\
Normalized mutual information is defined as\\
\begin{equation}
NMI(Y,L) =  \frac{I(Y,L)}{\sqrt{H(Y)H(L)}}
\end{equation}
where $Y$ and $L$ denote the ground truth labels and algorithm's output. $I(\cdot)$ is the mutual information which measures the information gain after knowing the partitions generated by algorithm. The entropy of $Y$ and $L$ are used for normalization purpose.\\
Purity is defined as\\
\begin{equation}
PUR(Y,L) = \frac{\sum_{i} \max_{j}\lvert L_{i}\cap Y_{j} \rvert}{n}
\end{equation}
where $L=\{L_{1},L_{2},...,L_{c}\}$ denotes the partition of clusters generated by algorithm and $Y = \{Y_{1},Y_{2},...,Y_{c}\}$ denotes the ground truth of clusters. Each cluster generated by algorithm is assigned to a real cluster which has the most same samples.

\begin{table*}
    \caption{Clustering results of various methods on ORL, AR, Umist, COIL20, COIL40, MNIST, and RCV1. For RCV1, some methods which need a long running time are ignored.  }
    \label{results}
    \begin{tabular}{ccccccccccccc}
				\toprule
				Dataset & Metric & SSC & LRR & SSCOMP & LRSC & SSRSC & LatLRR & DUDR & LSR & TRR & FLSR & FTRR \\
                \midrule
                \multirow{3}{*}{ORL} & ACC & 56.00 & 72.25 & 39.00 & 74.50 & 77.75 & 72.75 & 61.75 & 71.50 & 83.25 & 77.75 & \textbf{86.00} \\
                 & NMI & 70.06 & 83.42 & 58.95 & 83.47 & 86.53 & 85.81 & 73.83 & 82.57 & 91.11 & 86.61 & \textbf{91.51}\\
                 & PUR & 62.00 & 76.25 & 47.00 & 76.50 & 78.50 & 76.75 & 67.50 & 76.25 & \textbf{87.50} & 79.00 & 87.25 \\
                \midrule
				
                \multirow{3}{*}{AR} & ACC & 42.86 & 74.64 & 29.29 & 77.02 & 72.26 & 76.67 & 40.65 & 71.67 & 83.10 & 72.38 & \textbf{89.76}  \\
                 & NMI & 63.57 & 87.22 & 52.34 & 86.03 & 88.19 & 89.19 & 61.05 & 87.70 & 93.22 & 88.13 & \textbf{94.64}  \\
                 & PUR & 49.88 & 77.50 & 36.19 & 79.40 & 75.36 & 79.29 & 43.69 & 74.88 & 84.52 & 75.95 & \textbf{90.48} \\
                \midrule
				
                \multirow{3}{*}{COIL20} & ACC & 76.60 & 58.33 & 39.31 & 60.63 & 74.51 & 65.69 & 81.46 & 68.68 & 83.89 & 71.04 & \textbf{90.35}  \\
                 & NMI & 88.09 & 71.19 & 53.38 & 72.28 & 82.92  & 76.76 & 87.05 & 74.10 & 90.94 & 78.47 & \textbf{93.05} \\
                 & PUR & 83.47 & 61.67 & 78.26 & 62.22 & 77.50 & 69.65 & 83.06 & 70.90 & \textbf{92.85} & 75.07 & 91.04   \\
                \midrule

                \multirow{3}{*}{COIL40} & ACC & 63.13 & 60.42 & 18.92 & 58.23 & 57.01 & 60.52 & 69.13 & 56.88 & 78.37 & 59.20 & \textbf{78.58} \\
                 & NMI & 82.82 & 76.29 & 29.49 & 74.48 & 73.11 & 75.96 & 80.49 & 75.87 & 87.98 & 75.54 & \textbf{88.01} \\
                 & PUR & 72.05 & 62.81 & 23.16 & 64.24 & 60.94 & 64.24 & 75.10 & 62.74 & \textbf{85.31} & 62.27 & 81.56 \\
                \midrule

                \multirow{3}{*}{Umist} & ACC & 64.79 & 61.67 & 28.75 & 60.63 & 66.88 & 61.25 & 69.38 & 60.21 & 70.83 & 60.00 & \textbf{76.67} \\
                 & NMI & 75.38 & 72.95 & 41.11 & 72.08 & 75.14 & 70.27 & 77.42 & 71.58 & 77.86 & 70.82 & \textbf{85.09} \\
                 & PUR & 66.25 & 64.17 & 38.54 & 63.33 & 68.54 & 62.50 & 74.17 & 62.71 & 73.33 & 61.88 & \textbf{79.17} \\
                \midrule
                
                \multirow{3}{*}{MNIST} & ACC & 55.60 & 58.60 & 34.00 & 59.10 & 55.80 & 59.80 & 56.30 & 55.50 & 60.50 & 62.10 & \textbf{70.70} \\
                 & NMI & 50.14 & 54.69 & 32.72 & 52.75 & 52.74 & 55.50 & 47.94 & 54.96 & 56.34 & 52.31 & \textbf{66.72} \\
                 & PUR & 55.60 & 62.50 & 35.60 & 62.50 & 57.50 & 63.40 & 56.30 & 58.90 & 61.00 & 62.10 & \textbf{70.70} \\
                \midrule
                
               \multirow{3}{*}{RCV1} & ACC & - & - & 30.23 & - & - & - & - & 64.06 & 71.01 & 77.54 & \textbf{81.85}\\
                 & NMI & - & - & 2.86 & - & - & - & - & 42.05 & 48.66 & 54.89 & \textbf{59.66}\\
                 & PUR & - & - & 32.68 & - & - & - & - & 64.06 & 80.16 & \textbf{82.65} & 81.85\\
                \bottomrule
                
    \end{tabular}
\end{table*}

\begin{figure*}[!h]
\centering
\subfigure[t=1]{
\includegraphics[width=0.32\textwidth]{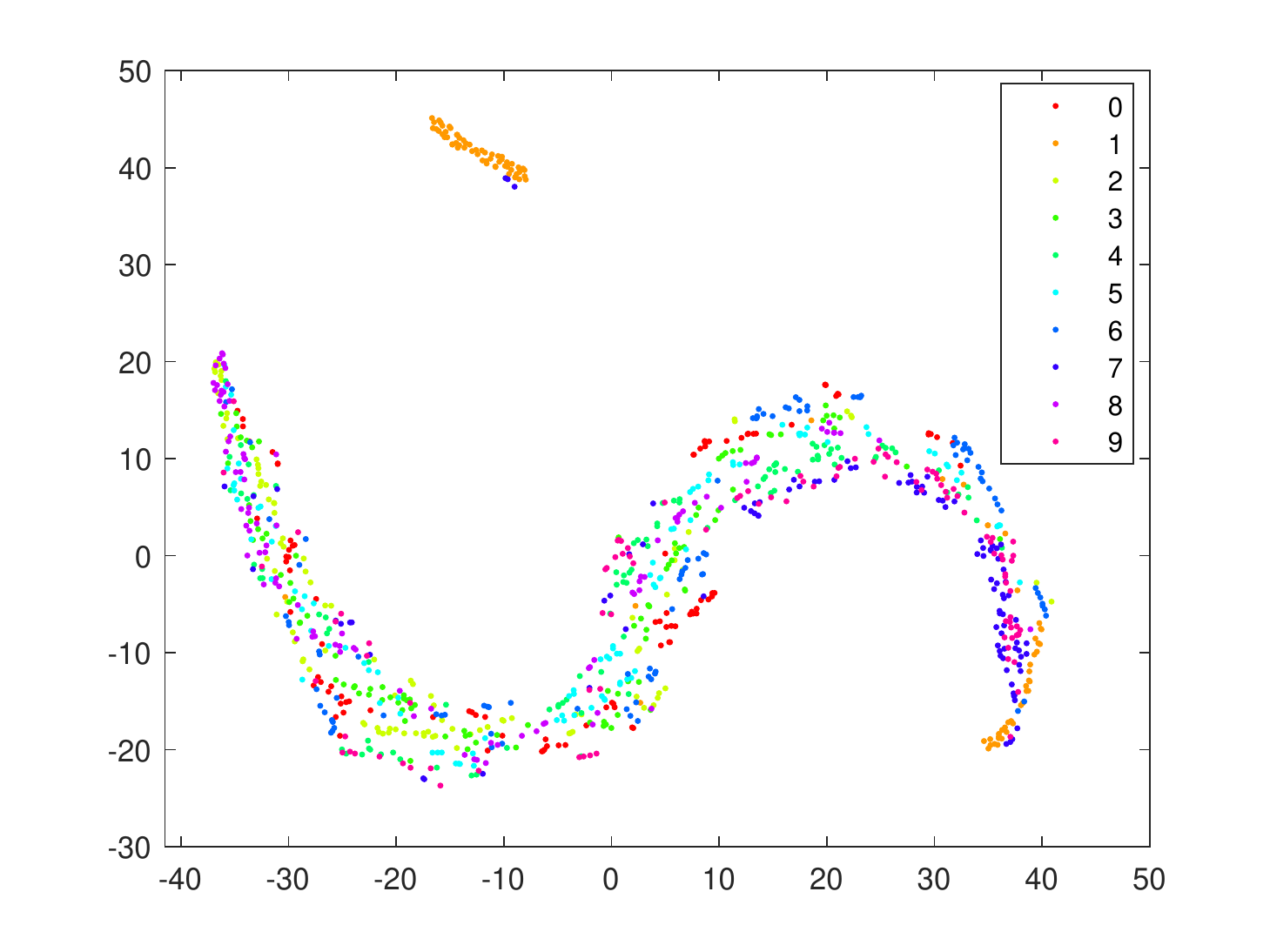}}
\subfigure[t=5]{
\includegraphics[width=0.32\textwidth]{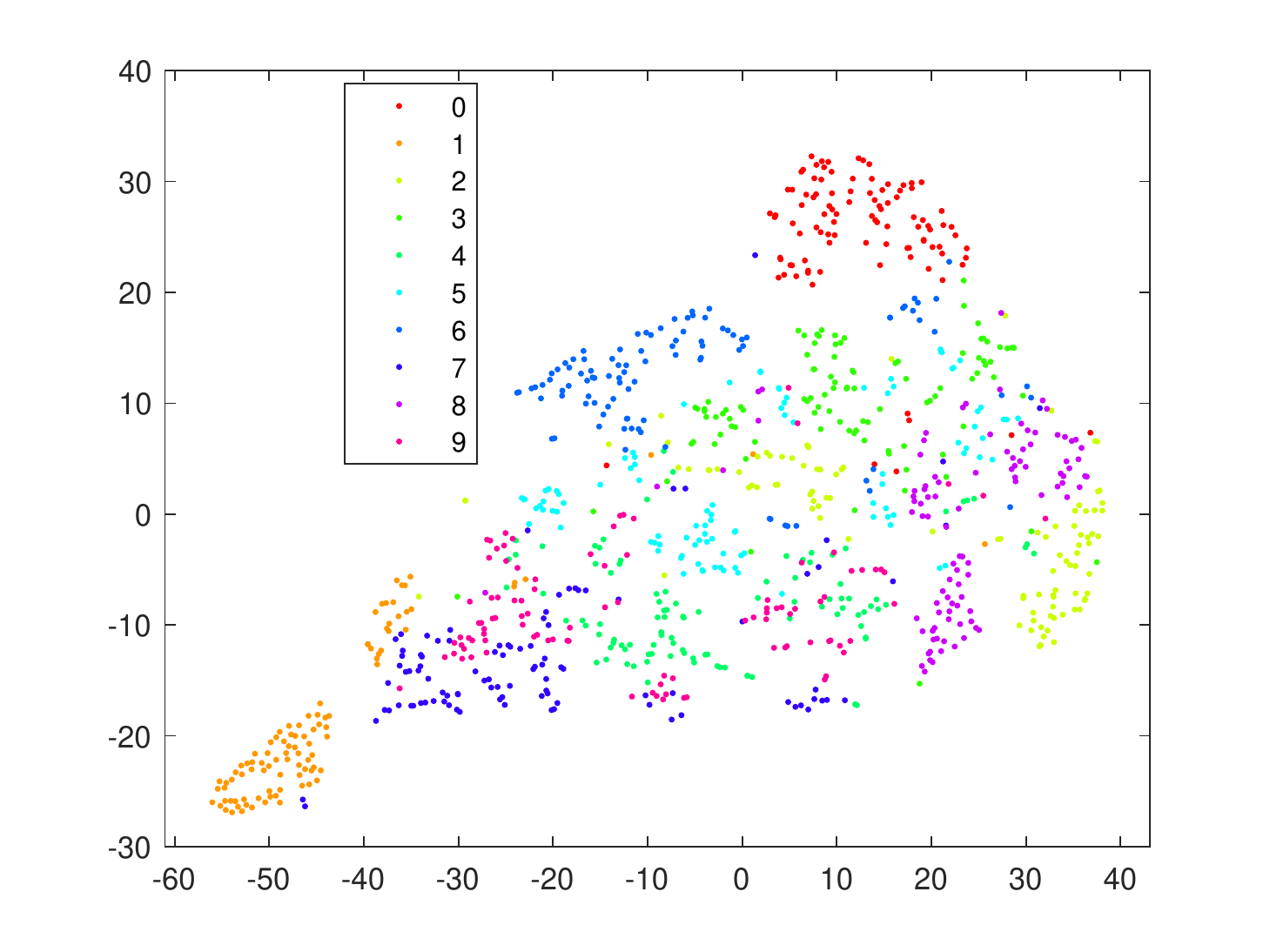}}
\subfigure[t=15]{
\includegraphics[width=0.32\textwidth]{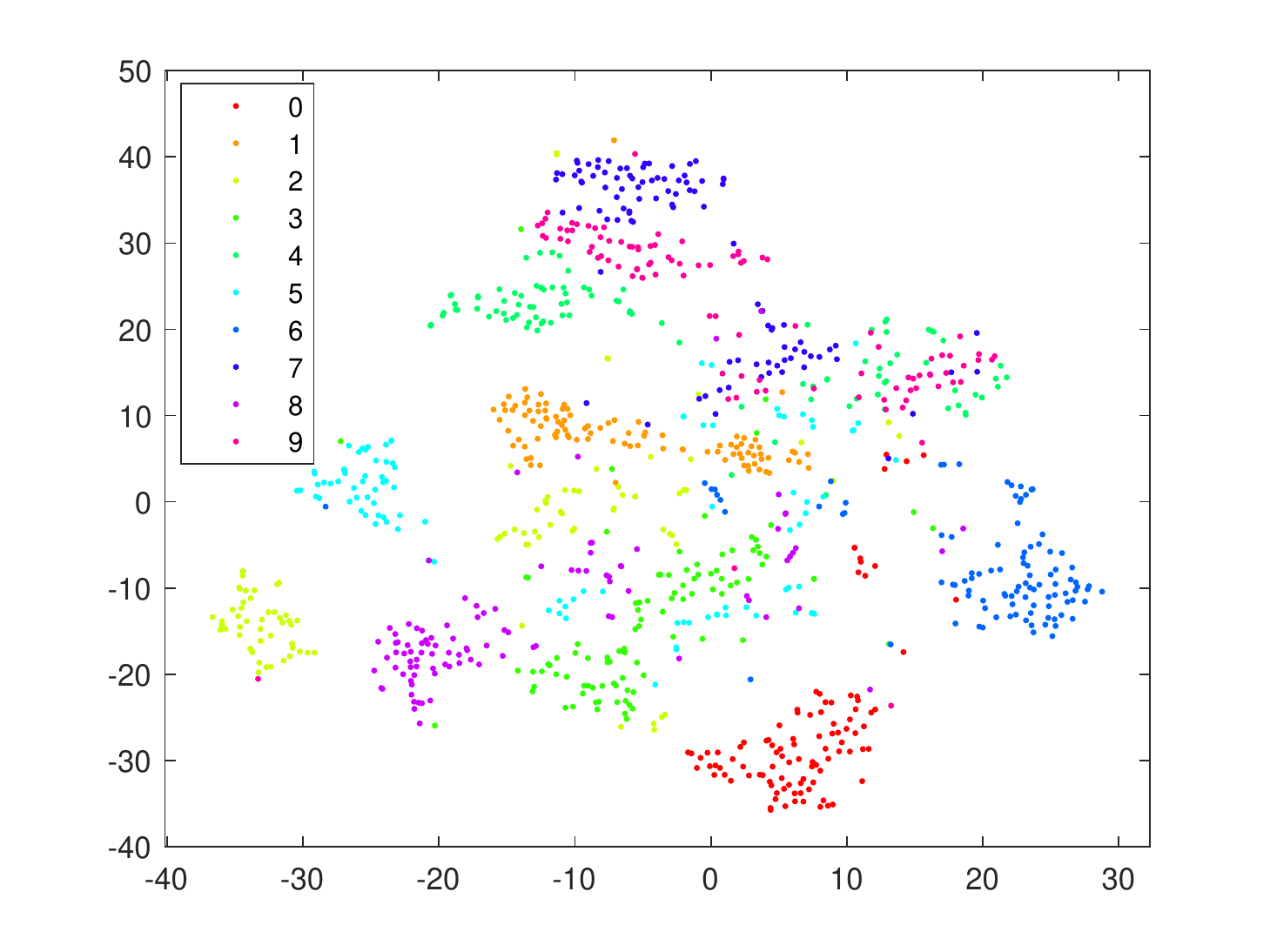}} % Reduce the figure size so that it is slightly narrower than the column.
\caption{t-SNE demonstration of filtered features of MNIST during the iteration process.}
\label{tsne}
\end{figure*}

\subsection{Results}
The results of experiments are summarized in Table \ref{results}, where the best results are highlighted in bold. We can observe that our purposed graph filtering framework boosts the performance of state-of-the-art subspace clustering techniques across most evaluation metrics. In particular,
\begin{itemize}
    \item FLSR and FTRR evidently improve the clustering results compared with the models they built upon. For example, FLSR and FTRR improve the accuracy of LSR and TRR by 4.47\% and 6.14\% respectively. This is attributed to the adoption of "clustering-friendly" representations realized by graph filtering. In particular, spatially close data points may help each other to prevent over-fitting in reconstructing the samples, i.e., the first term in Eq. (\ref{FLSR}). We use t-SNE to visualize the evolution of representation $\bar{X}$ as the iteration goes on in Fig.\ref{tsne}. As we can see, the filtered representation displays clearer cluster structure as the process goes on. The grouping effect of the filtered representation makes it much easier to separate the data points into disjoint subspaces.
    \item Compared to most recent method SRLSR, FTRR consistently outperforms it by a large margin. On average, accuracy, NMI, and purity are improved by 14.64\%, 10.07\%, and 13.64\%, respectively.
    \item FLSR clearly outperforms LatLRR on ORL, COIL20, and is comparable on COIL40 and Umist. Note that LatLRR performs learning in latent space,   which can extract salient features from hidden data, and thus can work much better than the benchmark methods that use the original data as features. Moreover, it demonstrates that LatLRR is more robust to noise with respect to dimension reduction based methods. Though LatLRR often performs better than LSR, its inferior to FLSR verifies that graph filtering is powerful.
    \item FTRR consistently outperforms LatLRR and DUDR by a very large margin. DUDR simultaneously performs dimension reduction and affinity graph construction. This shows that graph filtering could be more effective than dimension reduction approach in representation learning.
\end{itemize}

\begin{figure}[!htbp]
\centering
%\subfigure[FLSR]{\includegraphics[width=0.35\textwidth]{Affinity_Matrix_FSC1.pdf}}\\
\subfigure[TRR]{\includegraphics[width=0.4\textwidth]{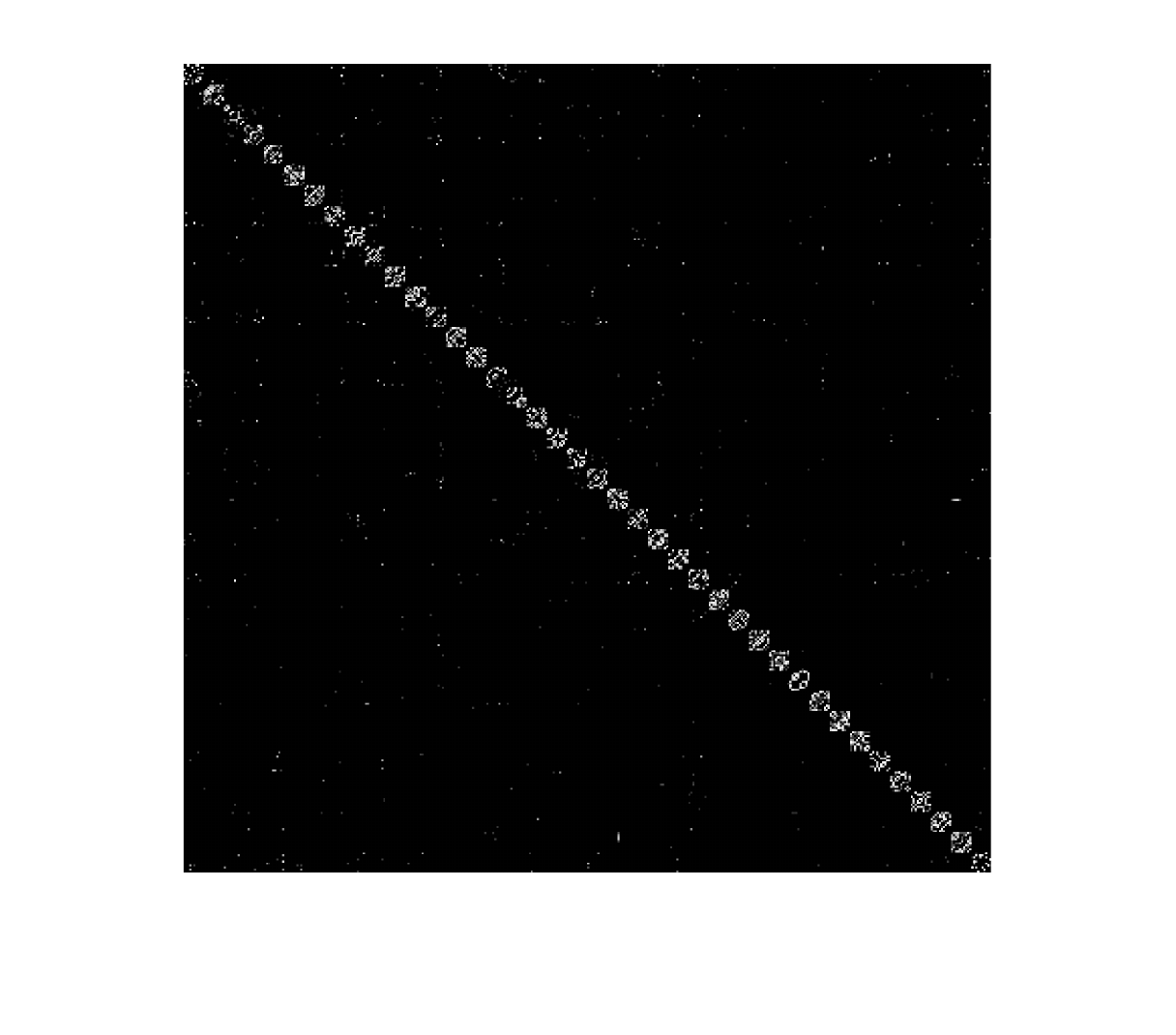}}
\subfigure[FTRR]{\includegraphics[width=0.4\textwidth]{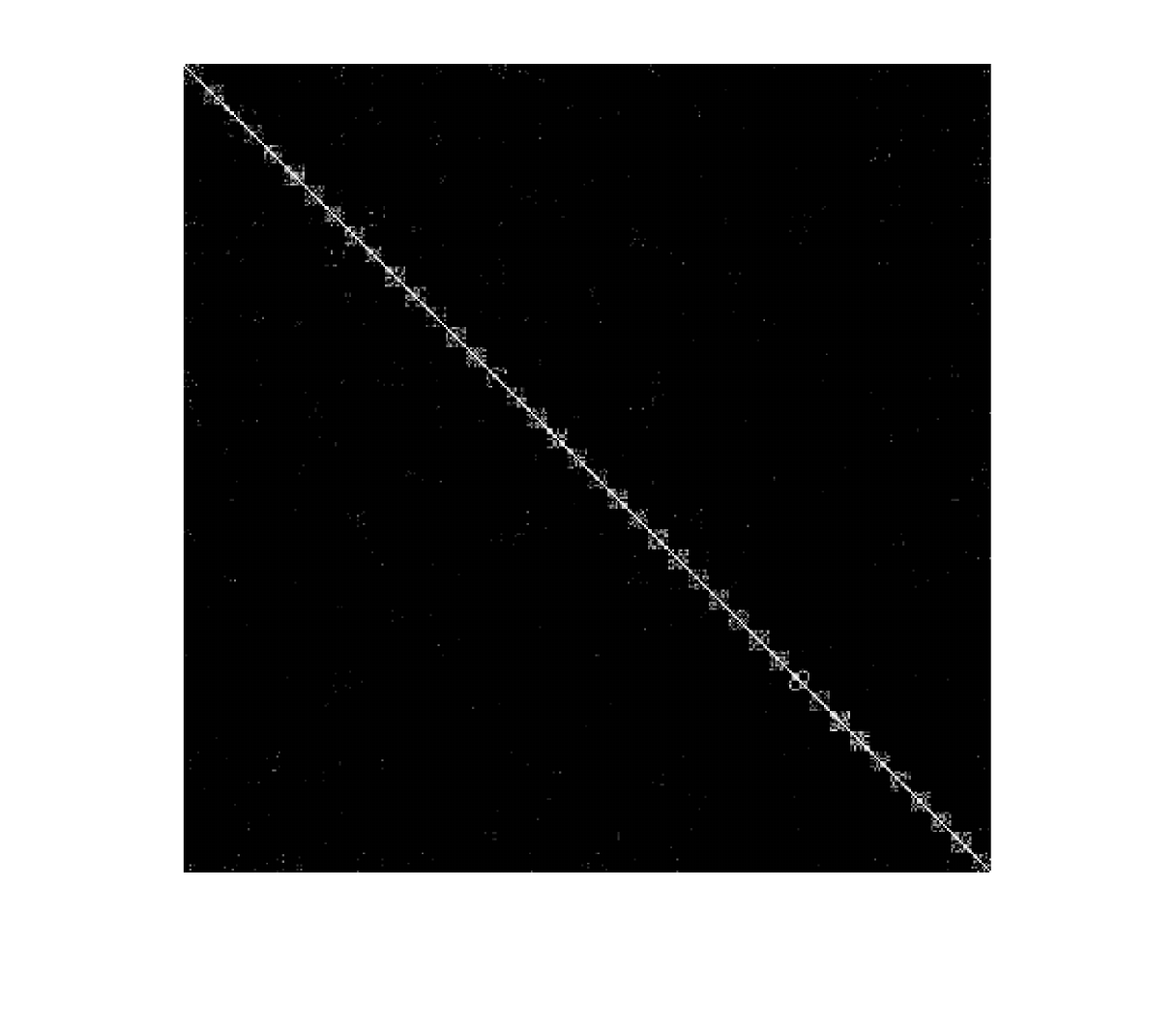}}\\
%\subfigure[LSR]{\includegraphics[width=0.35\textwidth]{Affinity_Matrix_LSR.pdf}}\\
 % Reduce the figure size so that it is slightly narrower than the column.
\caption{Affinity graph matrix obtained on ORL dataset.}
\label{Affinity Matrix}
\end{figure}

\begin{figure}[!htbp]
\centering
 %% label for secondsubfigure
\includegraphics[width=7cm]{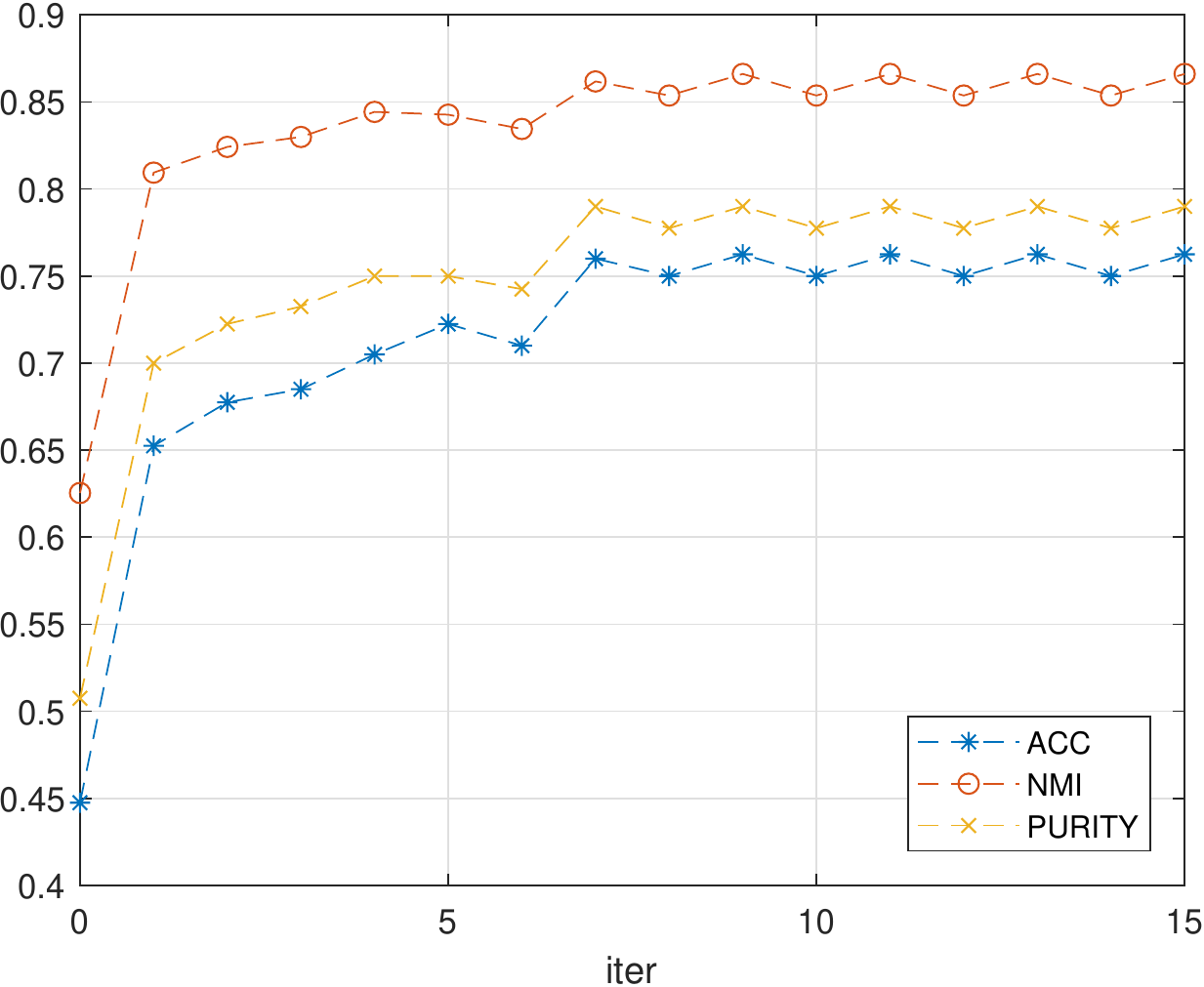}
\caption{The variation of performance throughout the iteration on ORL dataset. }
\label{evol}
\end{figure}
Take ORL as an example, we show the learned affinity graph matrix $W$ in Fig. \ref{Affinity Matrix}. Ideally, it should have a block-diagonal structure. As observed, both TRR and FTRR produce high-quality graphs. However, FTRR generates less noise than TRR. This explains why FTRR generates higher values in terms of accuracy, NMI, and purity.
%affinity matrices suffer from strong diagonal entries, indicating that for these two methods, the data points are mainly reconstructed by themselves.

In Fig. \ref{evol}, we also plot the change of accuracy, NMI, and purity values of FLSR as the iteration goes on. We can see that the performance increases quickly in the first 5 iterations and it reaches convergence after 7 iterations. Hence, we can see that our algorithm converges fast. Furthermore, the small fluctuations on the curve could be explained by the iterative nature of our algorithm.

\begin{figure*}[!htbp]
\centering
\includegraphics[width=0.33\textwidth]{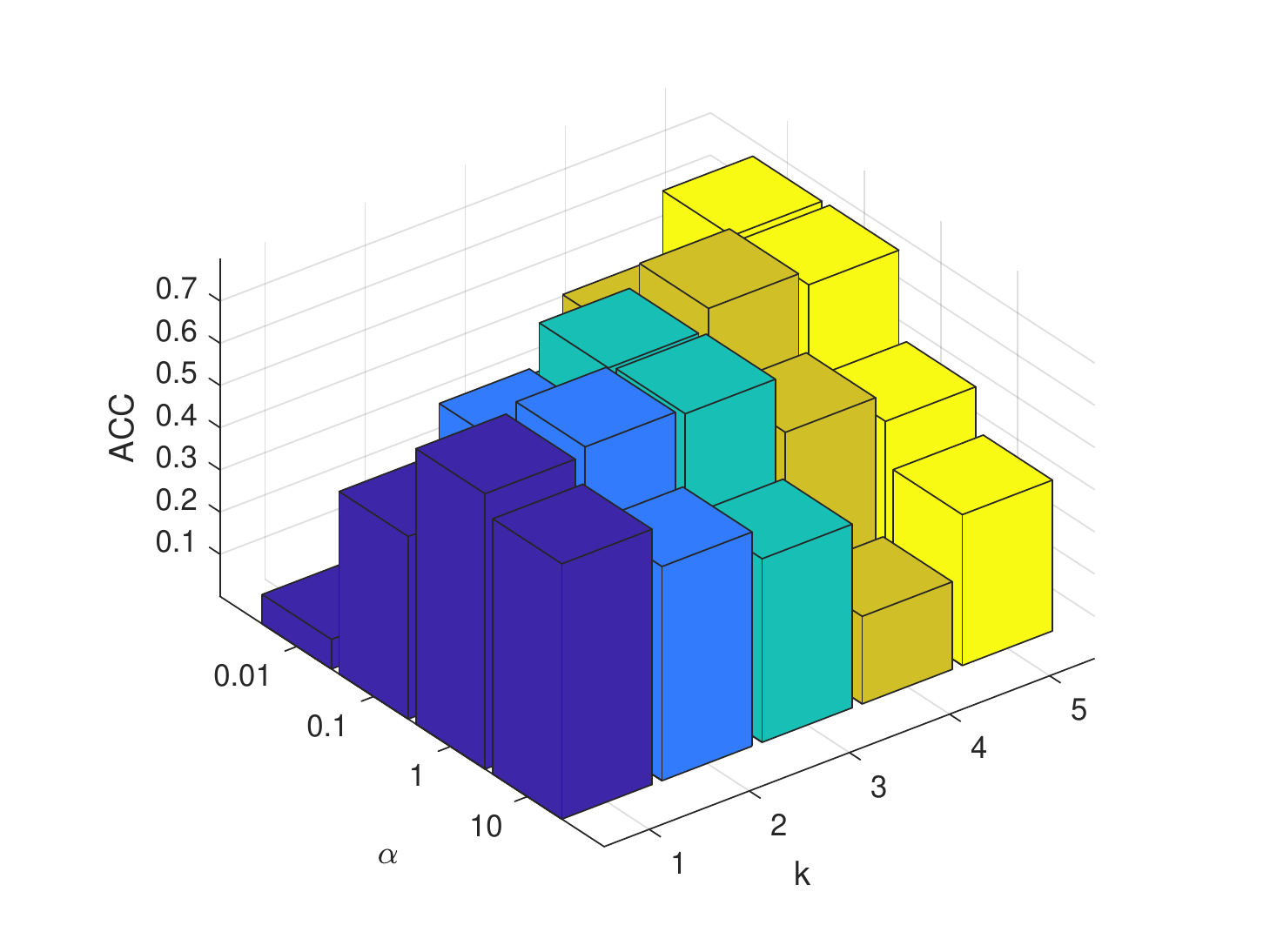}
\includegraphics[width=0.33\textwidth]{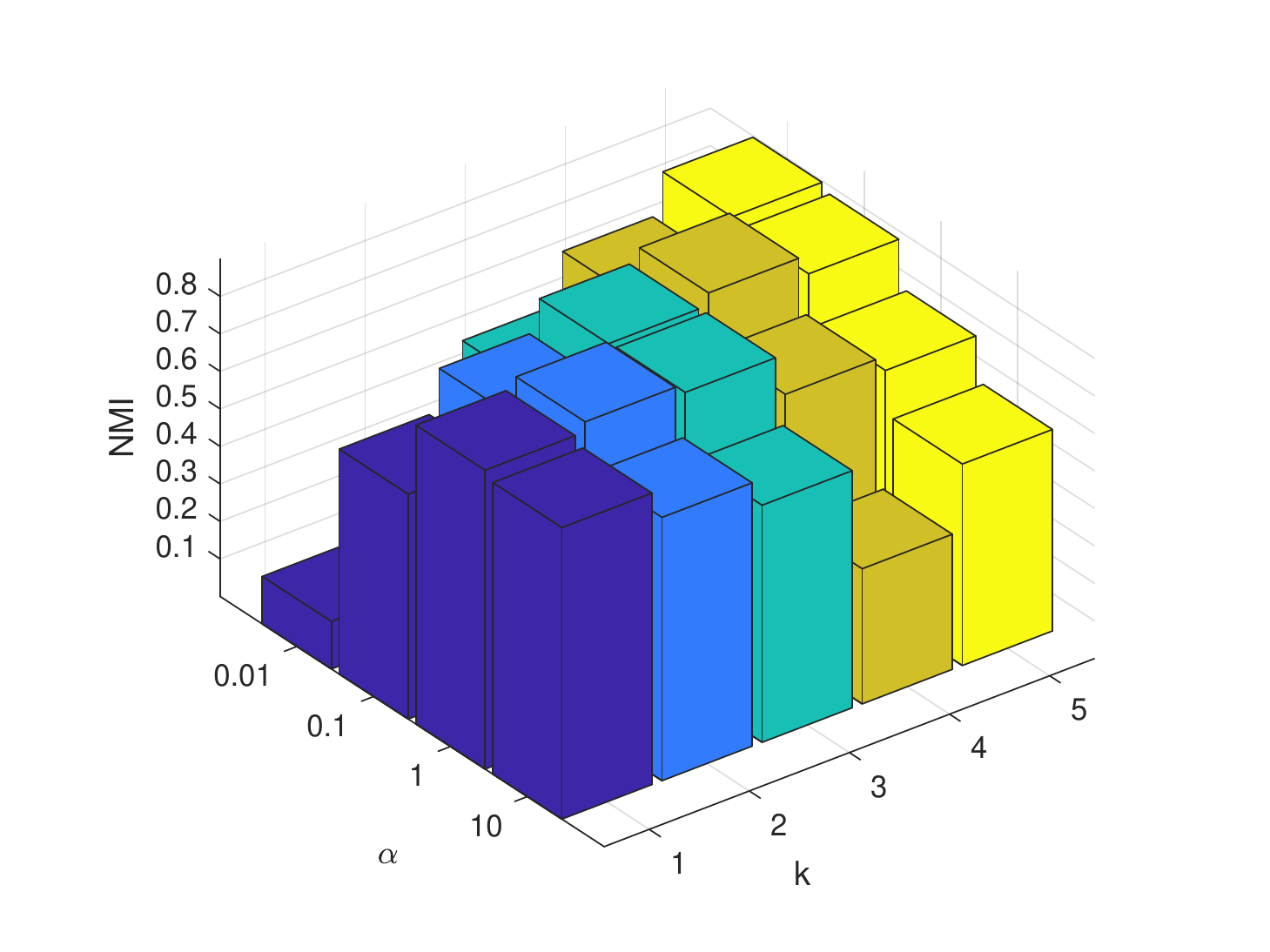}
\includegraphics[width=0.33\textwidth]{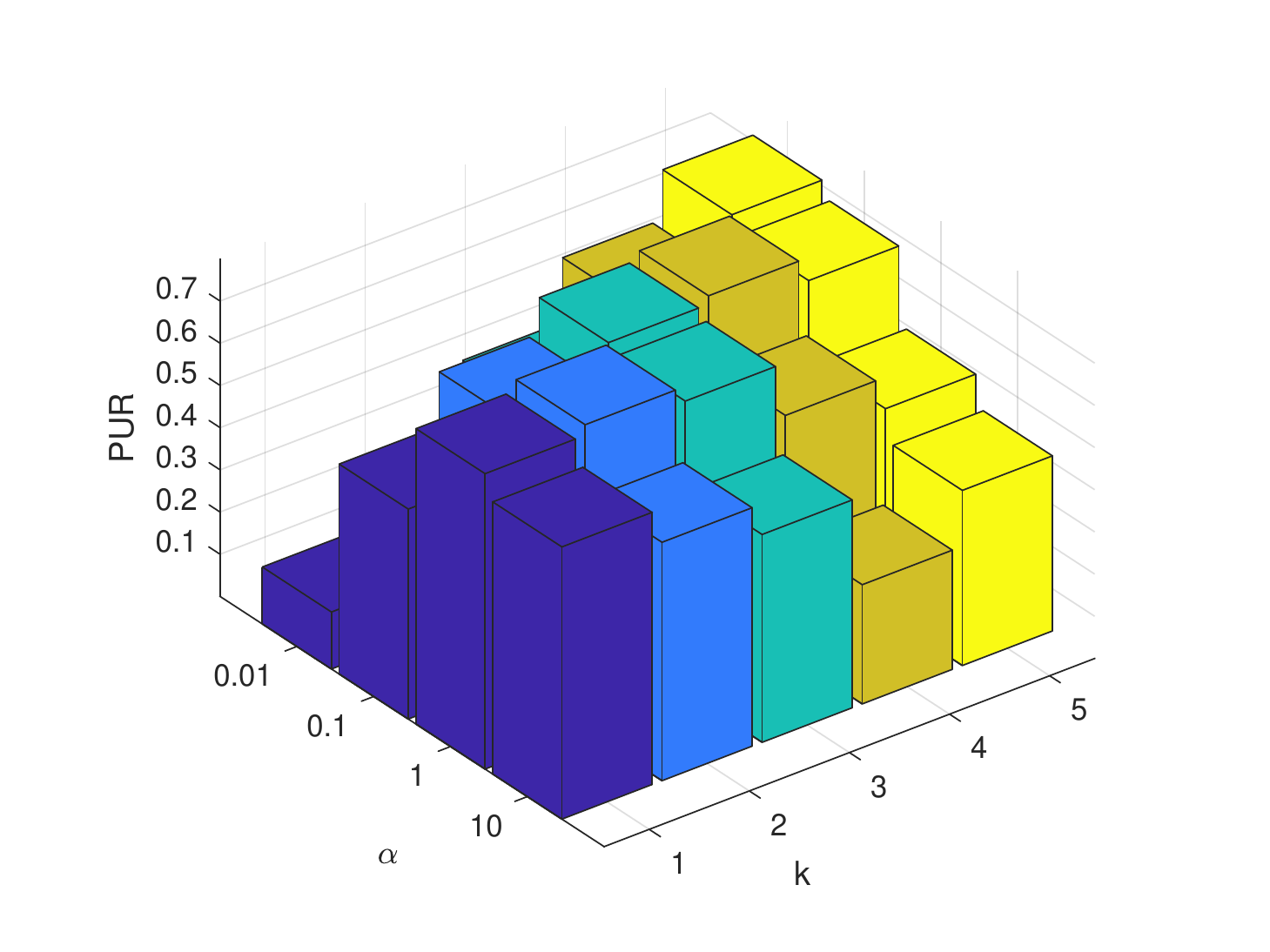} % Reduce the figure size so that it is slightly narrower than the column.
\caption{The influence of parameters $\alpha$ and $k$ for FLSR on ORL dataset.}
\label{ORL_FLSR}
\end{figure*}

\begin{figure*}[!htbp]
\centering
\includegraphics[width=0.33\textwidth]{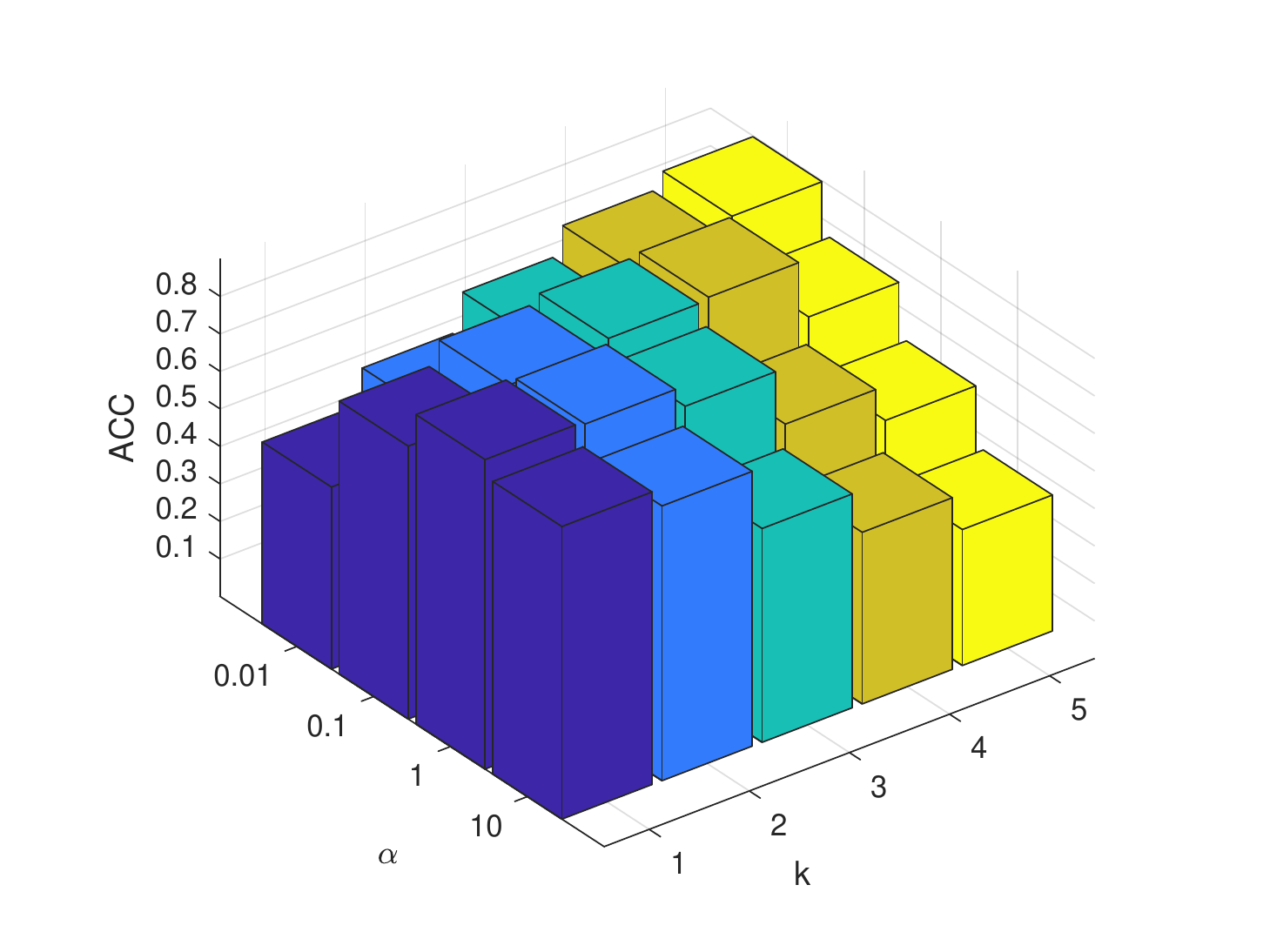}
\includegraphics[width=0.33\textwidth]{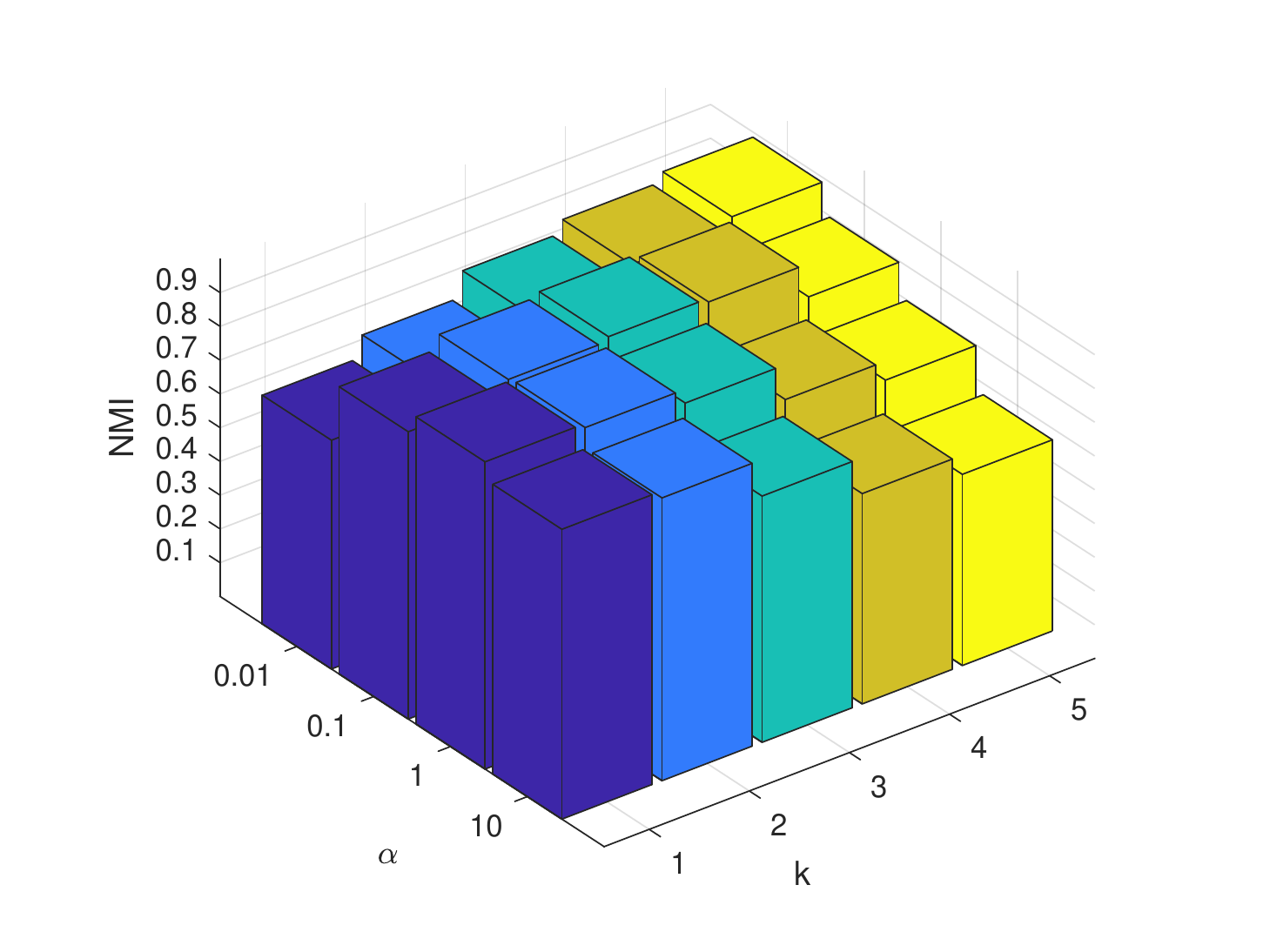}
\includegraphics[width=0.33\textwidth]{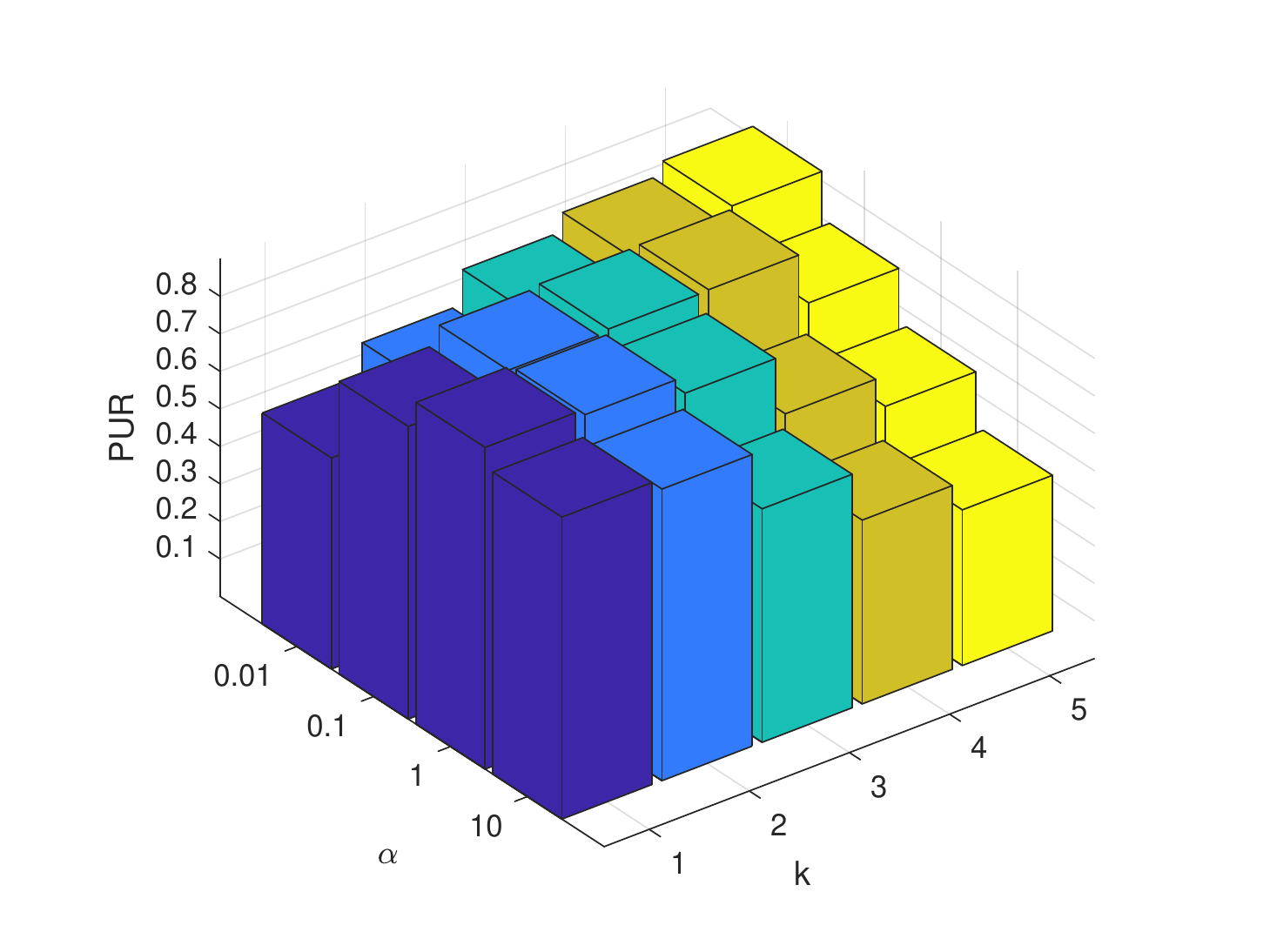} % Reduce the figure size so that it is slightly narrower than the column.
\caption{The influence of parameters $\alpha$ and $k$ for FTRR on ORL dataset.}
\label{ORL_FTRR}
\end{figure*}

%\begin{figure*}[htbp]
%\centering
%\includegraphics[width=0.33\textwidth]{ORL_acc_thresholding_fixed_6.pdf}
%\includegraphics[width=0.33\textwidth]{ORL_nmi_thresholding_fixed_6.pdf}
%\includegraphics[width=0.33\textwidth]{ORL_pur_thresholding_fixed_6.pdf} % Reduce the figure size so that it is slightly narrower than the column.
%\caption{The influence of parameters $\alpha$ and $k$ of FSC2 on ORL, while m fixed to 6.}
%\label{ORL_FSC2_m_fixed}
%\end{figure*}
 \setlength{\tabcolsep}{1.2pt}
\begin{table}
    \caption{Clustering results of FLSR and FTRR compared with deep methods.}
    \label{results:deep}
    \begin{tabular}{lrrrrrrrr}

				\toprule
				Dataset & Metric & AE-SSC & DSC-L1 & DEC & DKM &DCCM& FLSR & FTRR\\
                \midrule
                \multirow{3}{*}{ORL} & ACC & 75.63 & 85.50  & 51.75 & 46.82&60.00 & 77.75 & \textbf{86.00} \\
                 & NMI & 85.55 & 90.23 & 74.49 & 73.32&79.30 & 86.61 & \textbf{91.51}\\
                 & PUR & 79.50 & 85.85  & 54.00 & 47.52 &56.30& 79.00 & \textbf{87.25} \\
                \midrule
				
                \multirow{3}{*}{COIL20} & ACC & 87.11 & \textbf{93.14}  &72.15 & 66.51&81.40 & 71.04 & 90.35  \\
                 & NMI & 89.90 & \textbf{93.53}  & 80.07 & 79.71&87.10 & 78.47 & 93.05 \\
                 & PUR & 89.01 &\textbf{93.06}  & 69.31 & 69.64&80.10 & 75.07 &91.04   \\
                \midrule

                \multirow{3}{*}{COIL40} & ACC & 73.91 &\textbf{ 80.03} & 48.72 & 58.12&78.00 & 59.20 & 78.58 \\
                 & NMI & 83.18 & 88.52  & 74.17 & 78.40 &\textbf{89.10}& 75.54 & 88.01 \\
                 & PUR & 78.40 & \textbf{86.46}  & 41.63 & 63.67&77.10 & 62.27 & 81.56 \\
                \midrule

                \multirow{3}{*}{Umist} & ACC & 70.42 & 72.42  & 55.21 & 51.06&54.00 & 60.00 & \textbf{76.67} \\
                 & NMI & 75.15 & 75.56 & 71.25 & 82.49 &74.30& 70.82 & \textbf{85.09} \\
                 & PUR & 67.85 & 72.04  & 59.17 & 56.85 &58.30& 61.88 & \textbf{79.17} \\
                \midrule
                
                \multirow{3}{*}{MNIST} & ACC & 48.40 & \textbf{72.80}  & 61.20 & 53.32&42.50 & 62.10 &70.70 \\
                 & NMI & 53.37 & \textbf{72.17}  & 57.43 & 50.02&37.70 & 52.31 & 66.72 \\
                 & PUR & 52.90 & \textbf{78.90} & 63.20 & 56.47&45.00 & 62.10 & 70.70 \\
                
                \bottomrule
    \end{tabular}
\end{table}
\subsection{Parameter Analysis}
There is a trade-off parameter $\alpha$ in model (\ref{FLSR}). In addition, there is an implicit parameter $k$, i.e., the order of the filter. When $k$ increases, nearby node features become similar. However, too large $k$ will result in over-smoothing, i.e., the features of nodes in different clusters are mixed and become indistinguishable. Therefore, too large $k$ will deteriorate the clustering results. 

Taking ORL as an example, we show the effects of $\alpha$ and $k$ on clustering performance in Figs. \ref{ORL_FLSR} and \ref{ORL_FTRR}. We observe similar patterns on them. First, for a fixed $k$, the performance is enhanced when $\alpha$ increases. However, the performance is degraded when $\alpha$ has a large value. Second, it is easy to achieve good performance with a small $k$. Overall, a reasonable result can be achieved with a small range of $k$ and a large range of $\alpha$.

\subsection{Comparison with Deep Methods}
 Motivated by the success of deep neural networks (DNNs), unsupervised deep learning approaches are now widely used to learn nonlinear mappings from the data domain to low-dimensional latent spaces, which are supposed to be naturally suitable for clustering. Though our work is based on similar assumptions, we use the simple graph filtering technique instead. 
 
 We compare with some state-of-the-art deep clustering techniques, including SSC with pre-trained convolutional auto-encoder features (AE+SSC), Deep Subspace Clustering Network with L1-norm (DSC-L1) \cite{ji2017deep}, Deep Embedding Clustering (DEC) \cite{xie2016unsupervised}, and Deep K-means (DKM) \cite{fard2018deep}, Deep Comprehensive Correlation Mining (DCCM) \cite{wu2019deep}. In particular, DSC-L1 implements subspace clustering with DNNs and is closely related to our work. For a fair comparison, we directly copy the reported results for AE-SSC and DSC-L1 from \cite{ji2017deep,zhou2018deep}. For DEC, DKM, DCCM, we use the same encoder-decoder architecture in DEC \cite{xie2016unsupervised} and DKM \cite{fard2018deep}.

Table \ref{results:deep} shows the results given by various methods. We can observe the followings.\\
1) FTRR outperforms the strongest baseline DSC-L1 on ORL and Umist, and is inferior to DSC-L1 on COIL40, COIL20, and MNIST. Wilcoxon signed rank test gives a $p$ value of 0.56, thus FTRR and DSC-L1 are not statistically different from each other. Considering the complexity of training DNNs, graph filtering based subspace clustering is more appealing in practice. \\
2) FLSR generally outperforms DEC and DKM. FLSR also outperforms the most recent DCCM method on ORL, Umist, and MNIST. This shows that our learned representation is easy to cluster.\\
3) FTRR consistently outperforms DEC, DKM, and DCCM by a very large margin. This clearly demonstrate the effectiveness of graph filtering technique.

In summary, compared to deep clustering techniques, graph filtering approach is not only simple but also effective. From this perspective, our work falls into a family of recent efforts questioning the systematic use of complex deep learning methods without clear comparison to less fancy but simpler baselines \cite{dacrema2019we,lin2019neural}. Our approach can be a simpler alternative to deep clustering methods. 
\section{Ablation Study}

\begin{table*}
    \caption{The detailed analysis of the influence of filter order $k$ on PSNR, SSIM, and Fisher Score for noisy COIL40 data.}\label{psf}
    \label{results:deep}
    \begin{tabular}{lcccccccccccc}

				\toprule
				 Metric & Corrupted & $k=1$ & $k=2$ & $k=3$ & $k=4$ & $k=5$ & $k=6$ & $k=7$ & $k=8$ & $k=9$ & $k=10$\\
                \midrule
                PSNR & 26.12 & 28.76 & 28.30 & 27.76 & 27.34 & 27.02 & 26.75 & 26.52 & 26.32 & 26.15 & 25.99\\
                SSIM & 0.6995 & 0.8193 & 0.8435 & 0.8477 & 0.8477 & 0.8463 & 0.8444 & 0.8423	& 0.8400 & 0.8377 & 0.8354\\
                Fisher Score & $8.8\times10^4$	& $1.5\times10^6$ & $1.1\times10^7$ & $4.0\times10^7$ & $9.0\times10^7$ & $1.6\times10^8$ &	$1.6\times10^8$ &	$6.5\times10^7$ &	$4.0\times10^7$ &	$3.0\times10^7$ &	$2.2\times10^7$\\

                \bottomrule
    \end{tabular}
\end{table*}

\begin{table*}
    \caption{Clustering results on raw and filtered feature space of COIL40 dataset. }\label{diffres}%Row 1: Use raw data as input of filter;\\Row 2: Use noisy data (deviation=0.05) as input of filter. Affinity matrix is constructed by PKN.
    \label{results:deep}
    \begin{tabular}{lccccccccccccc}

				\toprule
				Data & Metric & Unfiltered & $k=1$ & $k=2$ & $k=3$ & $k=4$ & $k=5$ & $k=6$ & $k=7$ & $k=8$ & $k=9$ & $k=10$\\
                \midrule
                \multirow{3}{*}{Raw} & ACC & 77.81 & 82.19 & 91.04 & 91.22 & 92.85 & 92.99 & 91.67 & 92.26 & \textbf{93.40} & 93.09 & 92.33\\
                 & NMI & 88.20 & 90.19 & 95.41 & 95.15 & 96.41 & 96.35 & 96.42 & 96.48 & \textbf{96.97} & 96.75 & 96.44\\
                 & PUR & 81.67 & 84.20 & 91.98 & 91.70 & 93.33 & 93.06 & 93.26 & 93.40 & \textbf{94.51} & 94.20 & 93.44\\
                \midrule
				
                \multirow{3}{*}{Corrupted} & ACC & 71.32 & 86.29 & 90.90 & 91.91 & 91.81 & 92.01 & 92.01 & 92.08 & 93.06 & \textbf{93.13} & 92.05\\
                 & NMI & 84.81 & 92.78 & 95.78 & 95.90 & 95.94 & 96.24 & 96.15 & 96.20 & 96.69 & \textbf{96.80} & 96.01\\
                 & PUR & 77.57 & 87.81 & 92.33 & 92.40 & 92.88 & 93.13 & 93.13 & 93.00 & 94.27 & \textbf{94.27} & 92.92\\
                \bottomrule
    \end{tabular}
\end{table*}

Though we have demonstrated that graph filtering can improve the separability of raw data in Fig.\ref{tsne}, we further analyze this quantitatively. In particular, we also show that graph filtering has the effect of denoising, which also contributes to the performance improvement of downstream tasks. Taking COIL40 as an example, we add Gaussian noise with mean 1 and variance $\sigma=0.05$ to the raw features. Rather than using an iterative approach in our algorithm, we build the affinity graph using the probabilistic k-neatest method \cite{10.5555/3016100.3016174}. With this prior graph, it is easy to examine the effect of different orders of filter.

We first apply three metrics to systematically analyze the effect of filtering. Peak Signal to Noise Ratio (PSNR) is a standard measure for denoising, which relies strictly on numeric comparison. Structural Similarity Index (SSIM) \cite{2004ITIP...13..600W} is a popular metric to evaluate the structural similarities between images. Higher PSNR and SSIM indicate that the reconstruction is of higher quality with respect to the original images. Besides, we want to directly see how well graph filtering can separate samples in different clusters. Fisher Score \cite{fisher1936use} is a traditional metric to measure the linear separability of two sets of features. Thus, we can use it to evaluate the separability of data space before and after filtering. Therefore, higher Fisher Score indicates the feature space is more ``clustering-friendly".

For two clusters of samples $X^i$ and $X^j$, Fisher Score is calculated as the ratio of the variance between the classes (inter-class distance) to the variance within the classes (inner-class distance) under the best linear projection $\boldsymbol{w}$ of the original feature:
\begin{equation}
J\left(\boldsymbol{X}^{(i)}, \boldsymbol{X}^{(j)}\right)=\max _{\boldsymbol{w} \in \mathbb{R}^{m}} \frac{\left(\boldsymbol{w}^{\top}\left(\boldsymbol{\mu}^{(i)}-\boldsymbol{\mu}^{(j)}\right)\right)^{2}}{\boldsymbol{w}^{\top}\left(\boldsymbol{\Sigma}^{(i)}+\boldsymbol{\Sigma}^{(j)}\right) \boldsymbol{w}}
\label{fisher}
\end{equation}
where $\boldsymbol{\mu}^{(i)}$ and $\boldsymbol{\mu}^{(j)}$ represent the mean vector of $X^i$ and $X^j$ respectively, $\boldsymbol{\Sigma}^{(i)}$ and $\boldsymbol{\Sigma}^{(j)}$ represent the variance of $X^i$ and $X^j$ respectively. We can see that a larger $J$ indicates higher separability. It is known that the maximum separation occurs when $
\boldsymbol{w}=c\left(\boldsymbol{\Sigma}^{(i)}+\boldsymbol{\Sigma}^{(j)}\right)^{-1}\left(\boldsymbol{\mu}^{(i)}-\boldsymbol{\mu}^{(j)}\right)
$, where $c$ is a scalar. Plugging this into Eq.(\ref{fisher}), we obtain \\
$J\left(\boldsymbol{X}^{(i)}, \boldsymbol{X}^{(j)}\right)= \left(\boldsymbol{\mu}^{(i)}-\boldsymbol{\mu}^{(j)}\right)^{\top}\left(\boldsymbol{\Sigma}^{(i)}+\boldsymbol{\Sigma}^{(j)}\right)^{-1}\left(\boldsymbol{\mu}^{(i)}-\boldsymbol{\mu}^{(j)}\right)$. 

Then we perform graph filtering on both noisy data and raw data from $k=1$ to $k=10$. For each $k$, we compute the PSNR, SSIM, Fisher Score for each sample and report the average value in Table \ref{psf}. We can observe that PSNR, SSIM, and Fisher Score share the same trend, i.e., they increase when $k$ becomes larger at the beginning, then begin to drop when $k$ becomes too big. As mention earlier, this indicates that too large $k$ will incur over-smoothing. Specifically, PSNR jumps from 26.12 to 28.76 when we apply graph filter $k=1$ to the noisy images. For SSIM, it reaches its peak when $k=3$. These verify that graph filtering has the effect of removing noise and recovering the structure of images. This echos the findings in Fig.\ref{Affinity Matrix}. Fisher Score reaches its peak when $k=5$, which validates that graph filtering enhances separability of data representation. 

\begin{figure*}[!htbp]
\centering
\includegraphics{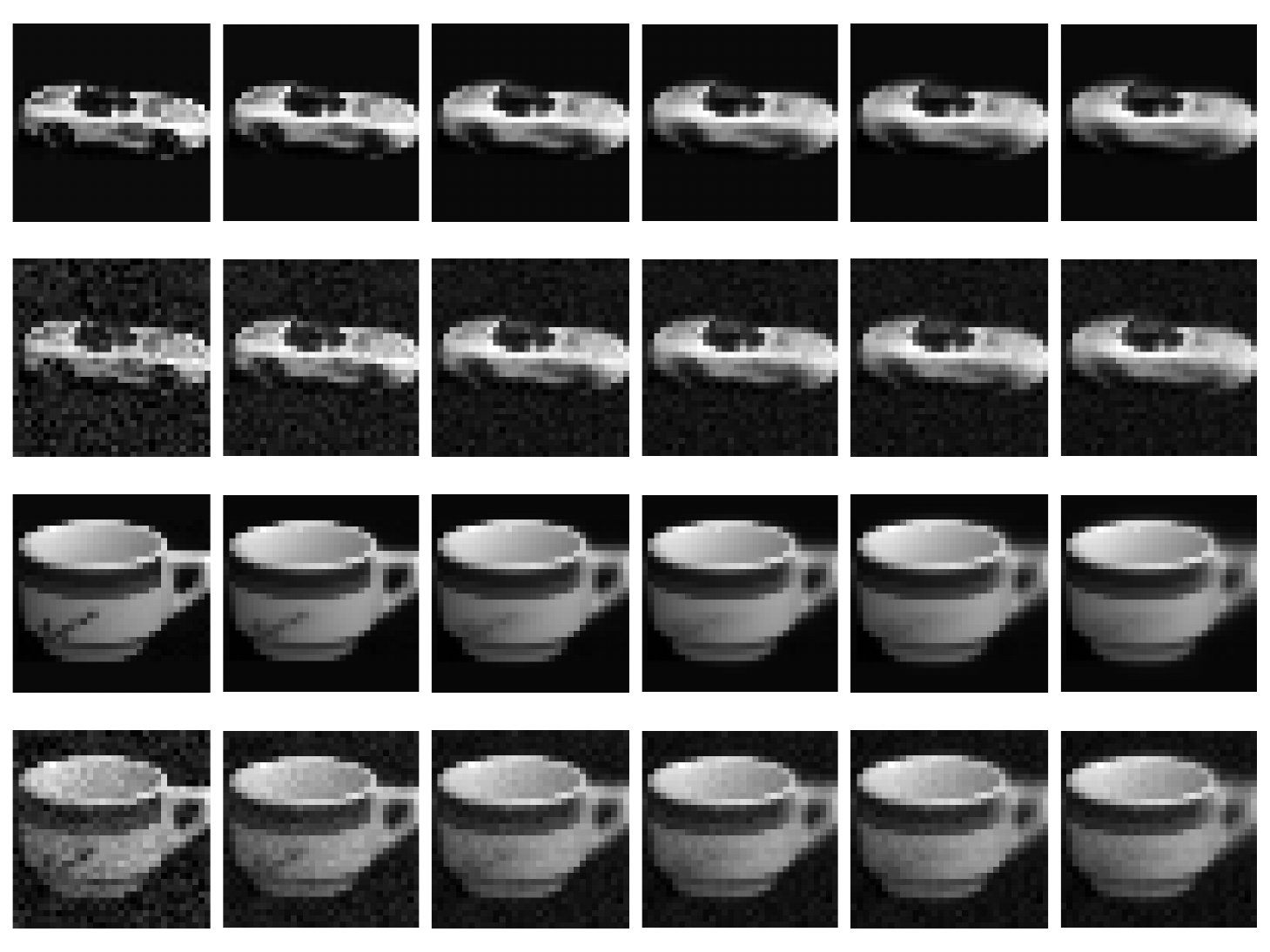}
\caption{Samples of images before and after graph filtering. Rows 1\&3 are two raw images while rows 2\&4 are their corresponding noisy images. From left to right, each column corresponds to filter order $k=$0 (unfiltered), 1, 3, 5, 7 and 10.}
\label{evovfig}
\end{figure*}

We visualize two sample images in Fig.\ref{evovfig}. We can observe that all images become smoother when the $k$ increases. From the second and fourth row, it can seen that noise is reduced from left to right. As a matter of fact, a natural image can be decomposed into a low spatial frequency component and a high spatial frequency part. The former contains the smoothly changing structure, e.g., background, and the latter one describes the rapidly changing fine details, e.g., outliers. This explains why our low-pass filter can help remove the noise. Too large $k$ will remove some specific properties of classes, which results in non-discriminative representations. This in turn makes it hard to separate those images, resulting in a decline in performance. 

Table \ref{diffres} further reports the clustering results under different $k$. Similarly, we can see that the performance increases until $k$ reaches 8 and 9 for raw and corrupted data, respectively. Though there is 6.5 gap on clustering accuracy between raw and corrupted data, we achieve similar performance after we introduce the graph filtering. Furthermore, the reported performance on COIL40 is much better than that in Tables \ref{results} and \ref{results:deep}. This is due to the fact that we use a different graph construction method \cite{10.5555/3016100.3016174}, which consequently produces high quality graph filtering. It has been proved that an ideal graph should consist of $g$-connected components \cite{kang2017twin}. \cite{10.5555/3016100.3016174} harnesses this nice property and proposes a rank constrained graph construction method. From this perspective, our performance in Table \ref{results} could be further improved if a prior graph with high-quality is available in advance. Of course, we must combine the best of both worlds. Without graph filtering, \cite{10.5555/3016100.3016174} generates clustering performance 0.8392, 0.9250, 0.8722, in terms of accuracy, NMI, purity, respectively. This is inferior to our performance.   
%In this section, we attempt to figure out the effectiveness on dealing with noisy data of filtering. Since no prior graph knowledge is given, we build the affinity matrix based on the probabilistic k-nearest method. Gaussian noise with $u=1$, $\sigma=0.05$ is added to the raw feature. Then, the low-pass filter with parameter order $k$ is constructed by affinity matrix, with noisy feature as input. Three metrics are applied to testing the effects of filtering. Peak Signal to Noise Ratio(PSNR) and Structural Similarity(SSIM)\cite{2004ITIP...13..600W} are used with the raw feature as reference. Higher PSNR and SSIM generally indicate that the reconstruction is of higher quality, which exposes the capability of filtering. In addition, Fisher-score\cite{wang2020demystifying} is to measure the linear separability of the feature space before and after filtering. Higher Fisher-score shows the feature space is more ``clustering-friendly".

\section{Conclusion}
In this paper, we make the first attempt to introduce graph filtering to subspace clustering. Our goal is to perform subspace clustering in a ``clustering-friendly" representation, i.e., the data representation displays cluster structure, which in turn facilitates the downstream clustering. This is realized by graph filtering technique. Since the graph is unavailable beforehand, we adopt an iterative strategy. Taking LSR and TRR benchmark as examples, we show the considerable improvements brought by the smooth representation on seven datasets. Moreover, we demonstrate that graph filtering approach reaches comparable performance as deep learning methods. An ablation study demonstrates that graph filtering can remove noise, preserve structure in the image, and increase the separability of classes. In the future, we can utilize the graph filtering in a wider scope, e.g., classification and semi-supervised learning.

\begin{acks}
This paper was in part supported by Grants from the Natural
Science Foundation of China (No. 61806045), the National Key R\&D Program of China (No. 2018YFC0807500), the Fundamental Research Fund for the Central Universities under Project ZYGX2019Z015, the Sichuan Science and Techology Program (Nos. 2020YFS0057, 2019YFG0202), the Ministry of Science and Technology of Sichuan Province Program (Nos. 2018GZDZX0048, 20ZDYF0343, 2018GZDZX0014, \\
2018GZDZX0034).
\end{acks}

\bibliographystyle{ACM-Reference-Format}
\bibliography{mm20}


\end{document}